\documentclass[10pt,twocolumn,letterpaper]{article}

\usepackage{cvpr}              

\usepackage{graphicx}
\usepackage{amsmath}
\usepackage{amssymb}
\usepackage{booktabs}

\usepackage{booktabs}
\usepackage{tabularx}

\usepackage{graphicx}

\usepackage{float}
\usepackage{multirow}
\usepackage{diagbox}

\usepackage{cuted}
\usepackage{capt-of}
\usepackage{pifont}
\usepackage{color}
\usepackage{bm}
\usepackage{makecell}
\usepackage{bbding}

\usepackage[mathscr]{eucal}

\usepackage[pagebackref,breaklinks,colorlinks]{hyperref}

\usepackage[capitalize]{cleveref}
\crefname{section}{Sec.}{Secs.}
\Crefname{section}{Section}{Sections}
\Crefname{table}{Table}{Tables}
\crefname{table}{Tab.}{Tabs.}

\def\cvprPaperID{7153} 
\def\confName{CVPR}
\def\confYear{2023}

\begin{document}

\title{Dual Diffusion Architecture for Fisheye Image Rectification: Synthetic-to-Real Generalization}


\author{Shangrong Yang, Chunyu Lin\thanks{Corresponding author: cylin@bjtu.edu.cn}, Kang Liao, Yao Zhao\\
Institute of Information Science, Beijing Jiaotong University\\
Beijing Key Laboratory of Advanced Information Science and Network, Beijing, 100044, China\\
{\tt\small \{sr\_yang, cylin, kang\_liao, cjzhang, yzhao\}@bjtu.edu.cn}
}

\maketitle

\begin{abstract}
Fisheye image rectification has a long-term unresolved issue with synthetic-to-real generalization. In most previous works, the model trained on the synthetic images obtains unsatisfactory performance on the real-world fisheye image. To this end, we propose a Dual Diffusion Architecture (DDA) for the fisheye rectification with a better generalization ability. The proposed DDA is simultaneously trained with paired synthetic fisheye images and unlabeled real fisheye images. By gradually introducing noises, the synthetic and real fisheye images can eventually develop into a consistent noise distribution, improving the generalization and achieving unlabeled real fisheye correction. The original image serves as the prior guidance in existing DDPMs (Denoising Diffusion Probabilistic Models). However, the non-negligible indeterminate relationship between the prior condition and the target affects the generation performance. Especially in the rectification task, the radial distortion can cause significant artifacts. Therefore, we provide an unsupervised one-pass network that produces a plausible new condition to strengthen guidance. This network can be regarded as an alternate scheme for fast producing reliable results without iterative inference. Compared with the state-of-the-art methods, our approach can reach superior performance in both synthetic and real fisheye image corrections.
\end{abstract}

\vspace{-1em}
\section{Introduction}

\begin{figure}[!t]
  \centering
  \includegraphics[scale=0.09]{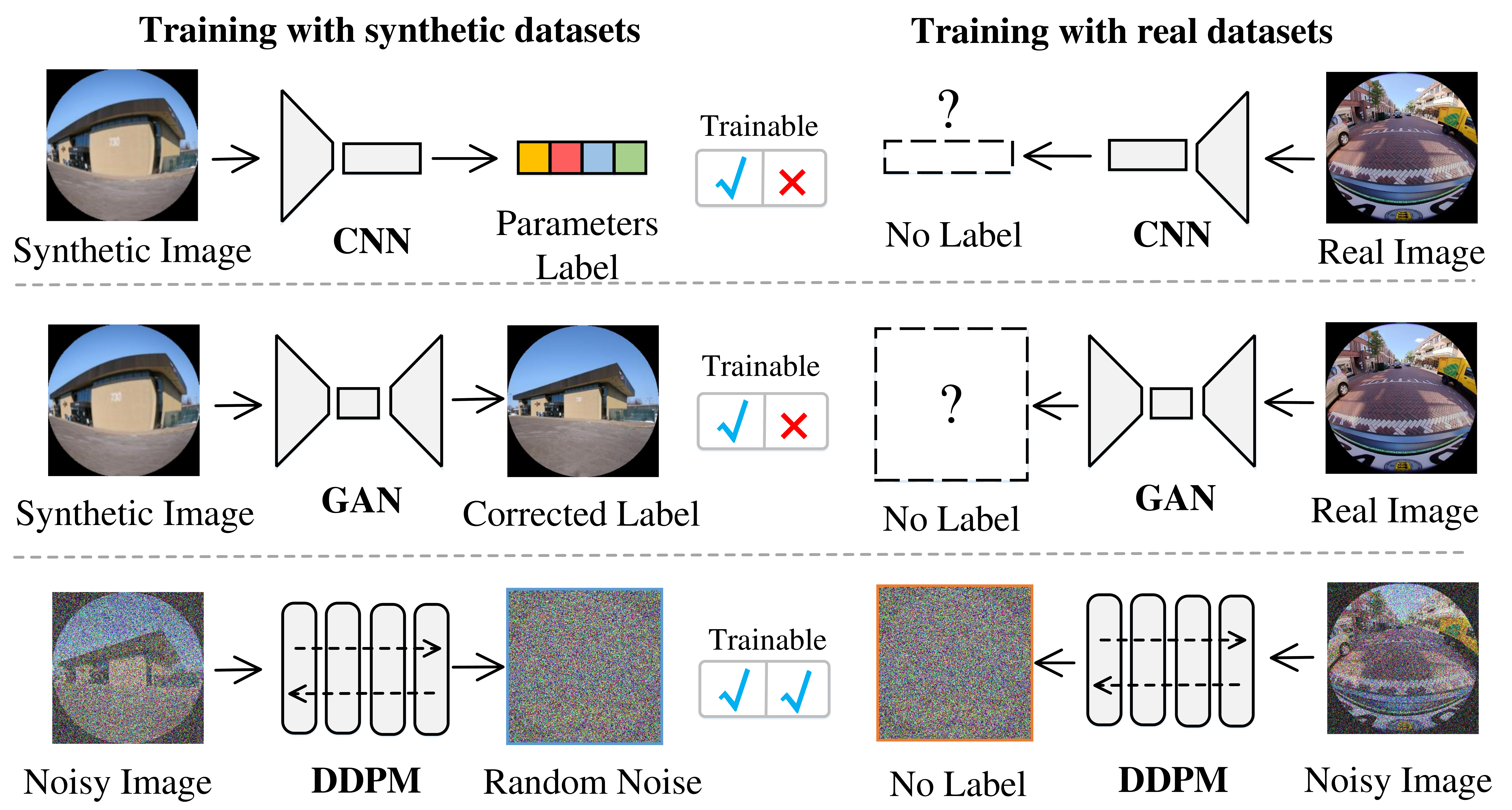}
  \caption{
  \label{three_network}
  CNNs and GANs both need labels for training. However, DDPMs can be trained by supervising the noise, which makes it possible to correct the real fisheye image.}
  \vspace{-0.5cm}
  \end{figure}
  
Many applications \cite{video_surveillance1}\cite{robot_navigation1}\cite{autonomous_driving1} have significant demands for large field-of-view environment information. Therefore, the fisheye camera is naturally taken into account. However, the images captured by fisheye cameras have structure distortion, which greatly affects the performance of subsequent vision algorithms \cite{Redmon2016YouOL}\cite{Shelhamer2017FullyCN}\cite{trajectory_prediction1}. To retain the performance of downstream tasks, one can consider correcting distorted images or redesigning subsequent algorithms. In general, many individuals prefer the simple former.

In the past, most calibration methods calculate distortion parameters by recognizing relevant features. Non-automatic calibration methods \cite{Reimer2013INTCAL1A}\cite{Reimer2009Intcal09AM}\cite{ZHANG1999} detect corners artificially using a checkerboard, while automatic methods \cite{Dansereau2013DecodingCA}\cite{Rui2014Unsupervised} use an algorithm that recognizes distinctive curves automatically. However, faulty detecting characteristics have a significant impact on these methods. As a result, neural networks are utilized to extract features in regard to their stable properties. \cite{Rong2016Radial}\cite{DeepCalib}\cite{Yin2018FishEyeRecNet}\cite{Xue2019} use deep regression models to predict distortion parameters. \cite{Liao2019}\cite{DDM}\cite{Blind} consider that it may be simpler to transform the correction into an image-to-image generation solution. By learning the complex empirical distributions, they can obtain the corrected results directly. Despite deep learning methods achieving significant advances in distortion correction, they can only train the network with synthetic datasets. However, there is a domain discrepancy between the synthetic and the real fisheye image. It raises the problem that the network has better performance on the synthetic fisheye images while obtaining worse performance on the real fisheye images.

The potential reason behind the poor testing effect on the real fisheye image is the unlabeled real fisheye image cannot be used for training. Existing convolutional neural networks (CNNs) and generative adversarial networks (GANs) both need paired image training, as shown in Figure \ref{three_network}. Therefore, we adopt a completely new framework, denoising diffusion probabilistic models (DDPM), to explore distortion correction. Existing DDPMs \cite{DDPM_super_resolution} \cite{DDPM_deblur} directly leverage the original image as the generating condition. However, the original and target image have non-negligible discrepancies in the fisheye rectification task, which greatly affects the generation quality. To address the disparities impact, we introduce a pre-correction network in the diffusion model and warp the fisheye images using the predicted distortion flow. The corrected image can be used as a more plausible condition for generating higher-quality results. It is worth mentioning that our pre-correction network is trained fully unsupervised. It can independently rectify fisheye images without the time-consuming inference in testing. 

\begin{figure}[!t]
  \centering
  \includegraphics[scale=0.0958]{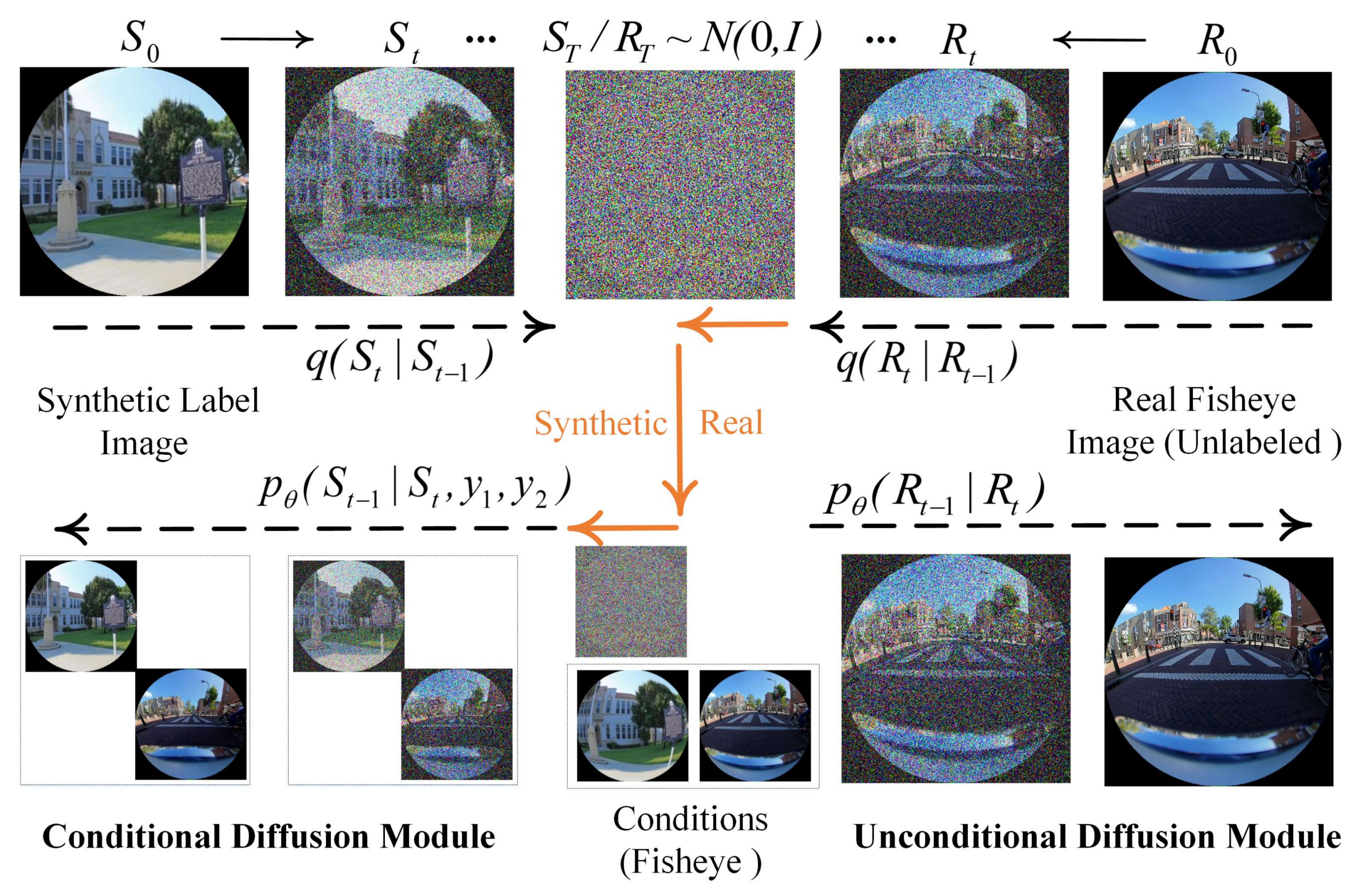}
  \caption{
  \label{forward_reverse}
  The generalization process in our dual diffusion architecture. By gradually introducing noises, the synthetic and real images can eventually develop into a consistent noise distribution, thus achieving the synthetic-to-real generalization correction.}
  \vspace{-0.5cm}
  \end{figure}

Because the diffusion model can be trained with noises, we further improve it and design a dual diffusion architecture. One part of our dual diffusion architecture is a conditional diffusion model, which employs paired synthetic datasets to learn the fundamental distribution of distortion. The other part is an unconditional diffusion model, which uses the unlabeled real fisheye image to facilitate the conditional diffusion model. To transform inconsistent data distribution into consistent noise distribution, we enforce the noise intensity of the two diffusion models to be consistent. Consequently, the network can achieve generalized distortion perception as well as cross-domain correction, as shown in Figure \ref{forward_reverse}. Thus, we no longer require the real fisheye image labels for training, but we can still leverage the conditional diffusion model to correct the real fisheye image during the testing stage.

The main contributions in this work are summarized as follows:

\begin{itemize}
  \item To mitigate the impact of data discrepancies, we generate a more acceptable condition from our incorporated unsupervised preprocessing network, which can also be used as an independent quick correction approach. 

  \item We exploit a dual-diffusion architecture that can convert inconsistent data distribution into consistent noise distribution to accomplish generalized distortion perception and rectification.

  \item Compared with previous methods, our method can introduce unlabeled real fisheye images for training. Our method achieves satisfactory results in both synthetic and real fisheye images.
\end{itemize}

\vspace{-0.3cm}
\section{Related Work}
The target of distortion correction is to restore the structure of the image before using downstream algorithms \cite{Lin2020FocalLF}\cite{tracking}\cite{Li2019TargetAwareDT}\cite{Girshick2014RichFH}\cite{Shelhamer2017FullyCN}. Early researchers \cite{Mei2007}\cite{Rui2014Unsupervised}\cite{Bukhari2013Automatic} \cite{Aleman2014Automatic} noticed that many straight lines under the conventional lens become curves in the fisheye perspective. Therefore, it is necessary to locate the feature corners or lines. Mei et al. \cite{Mei2007} developed a flexible calibration approach for calculating distortion parameters by locating corner points. However, it required additional standard planar grids and manual searching. Melo et al. \cite{Rui2014Unsupervised} designed an unsupervised calibration method. The automatic detection algorithm found a 'minimum of three lines' to achieve calibration. Although the automatic method \cite{Rui2014Unsupervised}\cite{Aleman2014Automatic} is more flexible than the manual method \cite{Mei2007}, feature detection was easily affected by image content, thereby hampering accurate calculation.

Many researchers employed reliable neural networks to tackle the problem of distortion. Rong et al. \cite{Rong2016Radial} first predicted several distortion intervals using convolutional neural networks (CNNs). Although a preliminary correction can be achieved, it is not complete owing to the simple network and quantization interval. \cite{Yin2018FishEyeRecNet}\cite{Xue2019} improved the regression network and introduced additional prior information, such as semantics and edges, for guidance. These methods significantly enhanced the correction performance. However, the image-parameters disparity limits the accurate prediction for all parameters. Therefore, \cite{Liao2019}\cite{PCN}\cite{Zhao2021RevisitingRD} introduced generation-based methods to immediately generate corrected images with learned empirical distributions. Liao et al. \cite{Liao2019} generated rectified results from the distribution learned by generative adversarial networks (GANs), but naive GANs led to visible artifacts. To further enhance the quality of the corrected image, \cite{DDM}\cite{PCN}\cite{Zhao2021RevisitingRD} proposed a multi-stage generation to separate the structure correction from the content reconstruction. By learning the corresponding relationship between two explicit distributions, the correction performance was boosted.

In practice, a large number of real fisheye images and corresponding labels are difficult to access. Therefore, the aforementioned deep-learning methods leveraged synthetic datasets for training. Since there is a domain discrepancy between the synthetic and the real datasets, the model trained on the synthetic images produces unacceptable results on the real images. Therefore, we design a dual diffusion model. With the help of denoising diffusion probabilistic models (DDPM) \cite{DDPM_1}\cite{DDPM_2}, we can train with both paired synthetic datasets and unlabeled real fisheye datasets. The same noise intensity is leveraged to supervise training. In this way, the model can learn similar distortion distribution and achieve cross-domain correction.

\section{Methodology}

\subsection{Fisheye Model}
The image was created by projecting coordinates in 3D space onto a 2D plane using a camera model. To capture as much information as possible with a larger FoV, the fisheye model change the pinhole model $d=ftan\theta $ to a nonlinear relationship between the incidence angle $\theta$ and the emergence angle $\rho $ :
\begin{equation}
  \rho=k_{1}\theta+k_{2}\theta^{3}+k_{3}\theta^{5}+\cdots
\end{equation}

However, this model involves precise angle calculation. Therefore, traditional methods \cite{Basu1995}\cite{Aleman2014Automatic}\cite{Bukhari2013Automatic}\cite{Barreto2005}\cite{Zhang2015}\cite{StraightLine}\cite{Rui2014Unsupervised} summarized two simple models: the polynomial model \cite{Basu1995} and the division model \cite{Aleman2014Automatic}. These models can neglect the angle and directly perform the coordinate transformation from the perspective image to the fisheye image. The principles of the polynomial model and the division model are similar, while the polynomial model does not need to handle the case where the denominator is 0. Therefore, we apply the polynomial model to synthesize the fisheye dataset. The polynomial model can be written as follows:
\begin{equation}
  \begin{bmatrix}x\\ y\end{bmatrix}=(1+\lambda_{1}{({r}')}^{2}+\lambda_{2}{({r}')}^{4}+\lambda_{3}{({r}')}^{6}+\cdots )\begin{bmatrix}{x}'\\ {y}'\end{bmatrix}
\end{equation}
Where the coefficient of the polynomial $\lambda_{n}$ is the distortion parameter. It reflects the distortion degree. $({x}',{y}')$ is an arbitrary point on the fisheye image, and its corresponding point on the perspective image is $(x,y)$. ${r}'$ is the distortion radius, which can be calculated by the Euclidean distance from $({x}',{y}')$ to the distortion center $(x_{d},y_{d})$. Similarly, the undistorted radius $r$ is the distance from $(x,y)$ to the image center $(x_{0},y_{0})$ on the perspective image.

\subsection{Diffusion Model}
DDPM \cite{DDPM_1}\cite{DDPM_2}\cite{DDPM_3}\cite{DDPM_4} are different from previous generation models \cite{GANs} \cite{VAEs} \cite{GLOW}. It breaks down an image generation task into several subtasks and includes a forward process $q$ with progressive noise addition and a reverse process $p$ with iterative noise removal.  Generally, its forward process \cite{DDPM_1}\cite{DDPM_3} can be represented as:
\begin{equation}
  q(x_{t}|x_{t-1})=\mathcal{N}(x_{t};\sqrt{1-\beta _{t}}x_{t-1},\beta _{t}\bm{I})
  \label{noisy_image}
\end{equation}

The next time data distribution $x_{t}$ can be calculated from the prior time distribution $x_{t-1}$. We can directly determine the distribution $x_{T}$ from initial distribution $x_{0} \sim q(x_{0})$ by iteratively multiplying $\prod_{t=1}^{T}q(x_{t}|x_{t-1})$. Therefore, the arbitrary distribution $x_{t}$ can be expressed as follows:
\begin{equation}
  q(x_{t}|x_{0})=\mathcal{N}(x_{t};\sqrt{\bar{\alpha }_{t}}x_{0},(1-\bar{\alpha }_{t})\bm{I})
\end{equation}

Where $\bar{\alpha _{t}}=\prod_{t=1}^{T}(1-\beta _{t})$. It should be emphasized that there are no trainable parameters in the forward process. Since the forward process constantly adds noise, the reverse process needs to denoise beginning with $x_{T} \sim \mathcal{N}(\bm{0},\bm{{\rm I}})$. In general, the reverse process can be written as neural network parameterization \cite{DDPM_vit}\cite{DDPM_semantic}:
\begin{equation}
  p_{\theta }(x_{t-1}|x_{t})=\mathcal{N}(x_{t-1};\mu _{\theta}(x_{t},t),\Sigma _{\theta }(x_{t},t))
\end{equation}

The purpose of network training is to minimize the distance $D$ between the forward and the backward distribution. $D$ can be calculated according to KL-divergence:
\begin{equation}
  D=D_{KL}(q(x_{t-1}|x_{t})||p_{\theta }(x_{t-1}|x_{t}))
\end{equation}

Under the hypothesis of Markov Chain, $q(x_{t-1}|x_{t})$ can be derived from $q(x_{t}|x_{t-1})$ :
\begin{equation}
  q(x_{t-1}|x_{t}, x_{0})=\mathcal{N}(x_{t-1};\tilde{\mu _{t}}(x_{t},x_{0}),\tilde{\beta  _{t}}\bm{I})
\end{equation}

Where $\tilde{\beta_{t}}=(\frac{1-\bar{\alpha}_{t-1}}{1-\bar{\alpha}_{t}})\beta_{t}$ and $\tilde{\mu _{t}}=\frac{1}{\sqrt{\alpha _{t}}}(x_{t}-\frac{\beta_{t}}{1-\bar{\alpha}_{t}}\epsilon _{t})$. Finally, the optimization function of the unconditional diffusion model can be written as follows:
\begin{equation}
  L_{u}(\theta )=\mathbb{E} \left \| \epsilon _{t}-\epsilon _{\theta }(\sqrt{\bar{\alpha }_{t}}x_{0}+\sqrt{1-\bar{\alpha }_{t}}\epsilon _{t},t)\right \|_{1}
\end{equation}

For the conditional diffusion model, we need to add additional condition $y$ to the network \cite{DDPM_synthesis}\cite{DDPM_super_resolution}\cite{DDPM_deblur} and replace $t$ with continuous noise level \cite{DDPM_generation}\cite{DDPM_super_resolution}\cite{DDPM_deblur}. Therefore, the optimization of the conditional diffusion model becomes:
\begin{equation}
  L_{c}(\theta )=\mathbb{E} \left \| \epsilon _{t}-\epsilon _{\theta }(\sqrt{\bar{\alpha }_{t}}x_{0}+\sqrt{1-\bar{\alpha }_{t}}\epsilon _{t},\bar{\alpha }_{t}, y)\right \|_{1}
\end{equation}

\begin{figure*}[!t]
  \centering
  \includegraphics[scale=0.075]{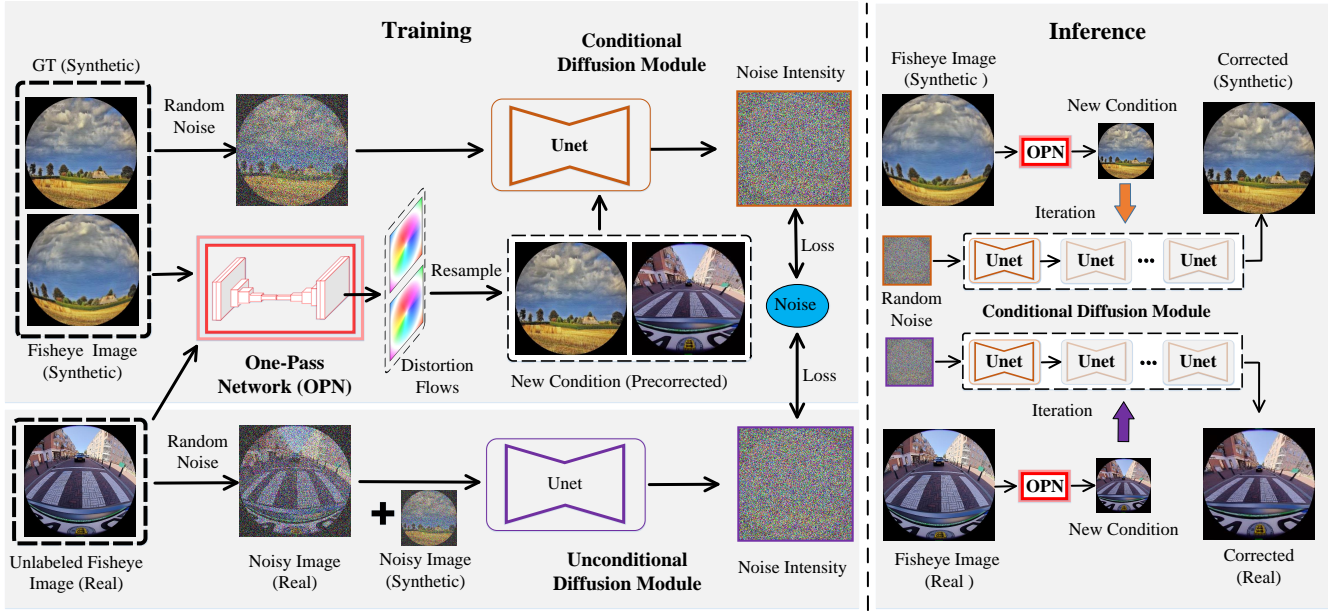}
  \vspace{-0.6cm}
  \caption{
  \label{structure}
  Our dual diffusion architecture (DDA) (left). It consists of three modules (conditional diffusion module, unconditional diffusion module, and one-pass network (OPN)) and trains on paired synthetic fisheye and unlabeled real fisheye datasets. For predicting the noisy intensity, the noisy synthetic ground truth and noisy real fisheye image are supplied to the conditional and unconditional diffusion modules, respectively. In the conditional diffusion module, a new reasonable guidance is generated from the one-pass network (OPN) correction. In the inference stage (right), the trained conditional diffusion module can gradually transform a random noise into a high-quality correction result by repeatedly predicting the noise intensity and recalculating new images.}
  \vspace{-0.3cm}
  \end{figure*}

\section{Architecture}
Since a large number of real fisheye labels is difficult to access, most existing generative-based correction methods \cite{DDM}\cite{PCN}\cite{Zhao2021RevisitingRD} can only utilize paired synthetic fisheye datasets for training. They have no opportunity to perceive the distribution of real fisheye images during the training phase. As a result, the network achieved well performance with synthetic fisheye images while showing blurred effects in real fisheye correction. Therefore, as shown in Figure \ref{structure}, we propose a dual diffusion architecture, which mainly includes three modules: conditional diffusion module, unconditional diffusion module, and one-pass network (OPN). Considering that we can access unlabeled real and paired synthetic fisheye images, we employ a conditional diffusion module to correct the synthetic fisheye images. Before the conditional correction, the OPN first predicts flows and generates coarse corrected images, which replace the original fisheye image as a new reasonable guidance. To perceive the real distribution, we simultaneously feed the unlabeled real fisheye image into an unconditional diffusion module and perform denoising. We add the same noise to the real and synthetic images, then transfer them between two diffusion modules to establish the relationship. Finally, our dual diffusion architecture can achieve training by supervising two same noise intensity.

\subsection{Conditional Diffusion Module}

Fisheye distortion correction is a conditional generation task like \cite{DDPM_deblur}\cite{DDPM_super_resolution}. Therefore, we synthesize the fisheye dataset and utilize the conditional diffusion model to perceive the fundamental distortion distribution. Our conditional diffusion module employs an Unet to forecast noise intensity ${\epsilon _{t}}'$ from the noisy perspective image $\bar{I}_{s}$. The encoder and decoder in Unet are ResNets with four scale outputs. The output channels are 64, 128, 256, and 512, respectively. The fundamental inputs of our module are noisy synthetic image $\bar{I}_{s}$ and conditional image $C_{s}$. $\bar{I}_{s}$ can be generated from the original perspective image ${I}_{s}$ with Formula \ref{noisy_image}. $C_{s}$ is the coarse correction result of the synthetic fisheye image $F_{s}$ in One-Pass Network (OPN), which will be mentioned later. In addition, we also correct the real fisheye image $F_{r}$ in the OPN. The result $C_{r}$ is concatenated with the fundamental inputs ($\bar{I}_{s}$, $C_{s}$) and sent to the conditional diffusion module. In contrast to the previous conditional DDPMs work \cite{DDPM_synthesis}\cite{DDPM_super_resolution}, our network incorporates two conditions as guidance, making it easier to simultaneously learn two distributions (synthetic and real) and accomplish domain generalization. Therefore, the optimization function of our conditional diffusion module is:
\begin{equation}
  L_{syn}=\mathbb{E} \left \| \epsilon -  S_{\theta}(\sqrt{\bar{\alpha}_{t}}I_{s}+\sqrt{1-\bar{\alpha}_{t}}\epsilon , \bar{\alpha}_{t},C_{s},C_{r}) \right \|_{1}
\end{equation}
where $S_{\theta }(\cdot )$ represents the conditional diffusion module.

\subsection{One-Pass Network (OPN)}
For generating the specific content, extra conditions should be included in existing conditional diffusion models. Generally, fisheye images $F_{s}$ can be used as the condition of the network. However, the fisheye image is the ineligible condition of its significant distortion. Other image generation tasks \cite{DDPM_super_resolution}\cite{DDPM_deblur} focus on enhancing the image quality, whereas distortion correction requires extra structure correction, which brings tremendous challenges to the network. To make the diffusion model concentrate on quality improvement, we introduce a one-pass network (OPN) to provide a new reasonable condition for the diffusion model. OPN is an encoder and decoder structure. Its encoder and decoder are composed of six convolutional layers. The output channels for each layer are 32, 32, 64, 128, 256, and 512, respectively. It takes the original condition image $F_{s}$ as input and produces a two-channel distorted flow $W$ \cite{PCN}, which reflects the image distortion degree. The distorted flow $W$ is then used to correct the original fisheye picture $F_{s}$ and generate the distortion-less corrected image $C_{s}$. Finally, we replace $F_{s}$ with $C_{s}$ as a new condition to assist the network predict the noise intensity ${\epsilon _{t}}'$. Because the distortion in $C_{s}$ has been greatly reduced, the network can predict the noise intensity more precisely and focus on quality improvement.

It is worth mentioning that our OPN is an entirely unsupervised network and does not require any labels for supervision. Furthermore, it offers an alternate fast correction scheme. Because of the distorted flow, we can immediately correct the fisheye images, thus avoiding the time-consuming inference in classic DDPMs. The experiment demonstrates the corrected results from OPN also have satisfactory performance as inference results.

\subsection{Unconditional Diffusion Module}
To achieve cross-domain perception, we need to introduce real fisheye images for training. However, there are no corresponding labels for real fisheye images. To solve this problem, we propose an unconditional diffusion module that is capable of label-free training to handle real fisheye images. It is equivalent to performing a general denoising task on a real fisheye image. There is no OPN in the unconditional diffusion module and its structure is the same as the conditional diffusion module. To establish information interaction, we add the coarse corrected result of real fisheye image $C_{r}$ as extra guidance to the conditional diffusion module and accept the noisy data $\bar{I}_{s}$ as an additional input to the unconditional diffusion module. To balance these two modules, we manually adjusted the noise intensity of the real fisheye image to match $\bar{I}_{s}$. In this way, despite the network having two noise data inputs, the unconditional diffusion module only has to predict one noise intensity for training. Therefore, the optimization target of the unconditional diffusion module can be represented as follows:
\begin{equation}
\begin{aligned}
    L_{real}=\mathbb{E} & \left \| \epsilon - R_{\theta}  (\sqrt{\bar{\alpha}_{t}}F_{r}+\sqrt{1-\bar{\alpha}_{t}}\epsilon, \right.
    \\
    &\phantom{=\;\;}\left. \sqrt{\bar{\alpha}_{t}}F_{s}+\sqrt{1-\bar{\alpha}_{t}}\epsilon, \bar{\alpha}_{t} ) \right \|_{1}
\end{aligned}
\end{equation}
where $R_{\theta }(\cdot )$ refers to unconditional diffusion network.

\begin{table}[htbp]
  \centering
    \begin{tabular}{m{8cm}}
    \toprule
    Algorithm 1 Training dual diffusion architecture \\
    \midrule
    Requirement: $S_{\theta}$, $R_{\theta}$, $OPN_{\theta}$ (conditional and unconditional diffusion module, one-pass network), $F_{s}$ \& $I_{s}$: paired synthetic images, $F_{r}$: real fisheye image \\
    1: repeat \\
    2: $C_{r}\leftarrow OPN_{\theta }(F_{r})$, $C_{s}\leftarrow OPN_{\theta }(F_{s})$ \\
    3: $t\sim Uniform\left \{ 1,...,T \right \}$ \\
    4: $\epsilon \sim \mathcal{N}(\bm{0},\bm{{\rm I}})$ \\
    5: Perform Gradient descent according\\
    \hspace{1cm} $ \bigtriangledown_{\theta } || 2\epsilon -  
    S_{\theta}(\sqrt{\bar{\alpha}_{t}}I_{s}+\sqrt{1-\bar{\alpha}_{t}}\epsilon , \bar{\alpha}_{t},C_{s},C_{r})  - $ \\ 
    \hspace{0.8cm} $R_{\theta}(\sqrt{\bar{\alpha}_{t}}F_{r}+\sqrt{1-\bar{\alpha}_{t}}\epsilon, \sqrt{\bar{\alpha}_{t}}I_{s}+\sqrt{1-\bar{\alpha}_{t}}\epsilon, \bar{\alpha}_{t} )||_{1}$\\
    6: until converged \\
    \bottomrule
    \end{tabular}%
    \label{training_stage}
\end{table}%
\begin{table}[htbp]
  \centering
    \begin{tabular}{m{8cm}}
    \toprule
    Algorithm 2 Correct the synthetic and real fisheye image \\
    \midrule
    1: $C_{r}\leftarrow OPN_{\theta }(F_{r})$, $C_{s}\leftarrow OPN_{\theta }(F_{s})$ \\
    \hspace{3.7cm} $\triangleright $ One-pass correction scheme\\
    2: $C_{r(T)}, C_{s(T)} \sim \mathcal{N}(\bm{0},\bm{{\rm I}}) $  \\
    3: for $t=T,...,1$, do                         \\
    4: \hspace{0.2cm} $t\sim Uniform\left \{ 1,...,T \right \}$ \\
    5: \hspace{0.2cm} $z \sim \mathcal{N}(\bm{0},\bm{{\rm I}})$ if $t>1$, else $z=0$ \\
    6: \hspace{0.2cm} $C_{r(t-1)}=\frac{1}{\sqrt{\alpha _{t}}}(C_{r(t)}-\frac{\beta _{t}}{\sqrt{1-\bar{\alpha}_{t}}}S_{\theta }(\tilde{s_{t}},\bar{\alpha}_{t},\tilde{s_{t}},C_{r}))$                         \\
    \hspace{0.2cm} \hspace{0.3cm} $C_{f(t-1)}=\frac{1}{\sqrt{\alpha _{t}}}(C_{f(t)}-\frac{\beta _{t}}{\sqrt{1-\bar{\alpha}_{t}}}S_{\theta }(\tilde{s_{t}},\bar{\alpha}_{t},\tilde{s_{t}},C_{s}))$ \\
    \hspace{3.7cm} $\triangleright $ Inference correction scheme\\
    7: end for \\
    8: return $C_{r}$, $C_{s}$, $C_{r(0)}$, $C_{s(0)}$\\
    \bottomrule
    \end{tabular}%
\label{inference_stage}%
\vspace{-0.5cm}
\end{table}%

\subsection{Training strategy}
Benefiting from the DDPMs, our dual diffusion architecture can complete training by simply supervising the predicted noise, as shown in Algorithm 
\textcolor{red}{1}. Since our network is composed of conditional and unconditional diffusion modules, we need to optimize the two modules jointly. The final loss function is:
\begin{equation}
  L=L_{syn}+L_{real}
\end{equation}

Through the integrated supervision of the two modules' noises, our network can achieve end-to-end training.

\subsection{Inference Process}
Our network simultaneously uses the synthetic and the real fisheye image for training. Therefore, we can correct both at the same time. As shown in Algorithm 
\textcolor{red}{2}, we first use the OPN to predict the distortion flow $W$ and correct the synthetic and real fish images. Then we randomly sample a noise $\epsilon \sim \mathcal{N}(\bm{0},\bm{{\rm I}})$, which is concatenated with the synthetic or real corrected result and sent to the conditional diffusion module to predict the noise intensity. By repeatedly predicting the noise intensity and recalculating new images, we can obtain high-quality results.

\begin{figure*}[!t]
  \centering
  \includegraphics[scale=.17]{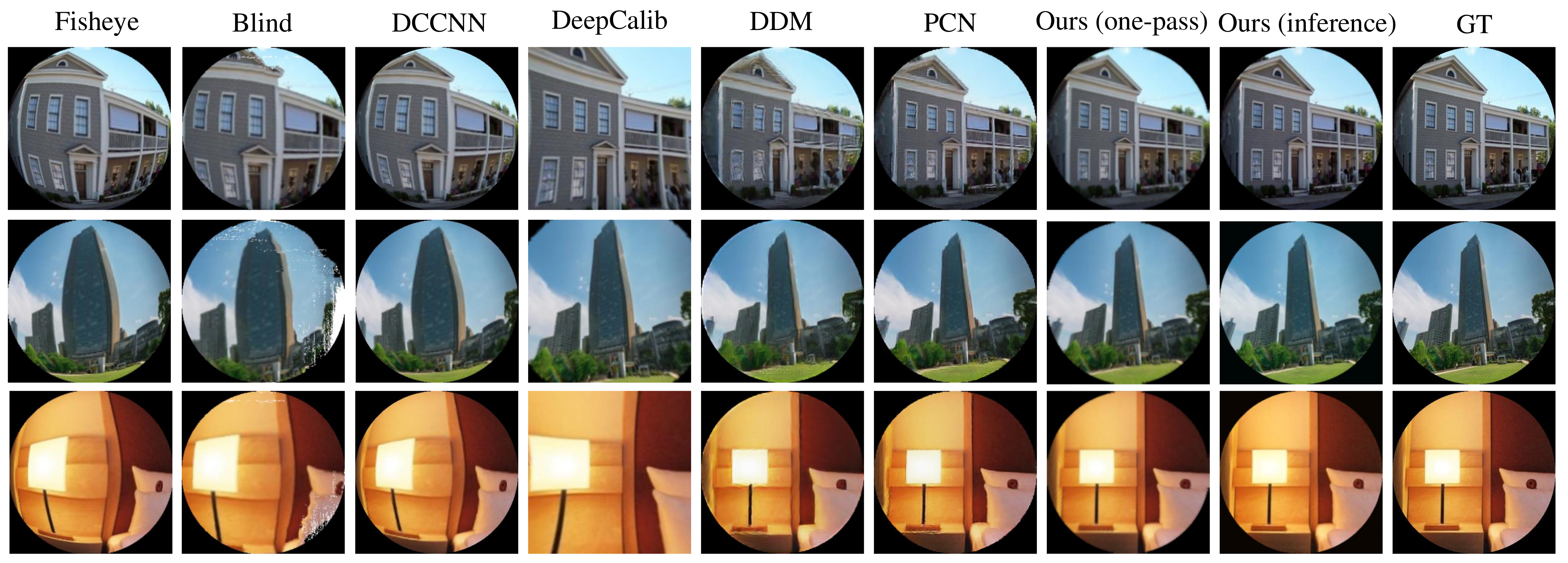}     
  \vspace{-0.2cm}
  \caption{
  \label{result}
  \textbf{Subjective comparison results on synthetic images.} We generate synthetic fisheye images with random distortion and test them using the state-of-the-art methods (Blind \cite{Blind}, DCCNN \cite{Rong2016Radial}, DeepCalib \cite{DeepCalib}, DDM \cite{DDM}, PCN \cite{PCN}) and our own.}
  \vspace{-0.2cm}
  \end{figure*}

\begin{figure*}[!t]
  \centering
  \includegraphics[scale=.17]{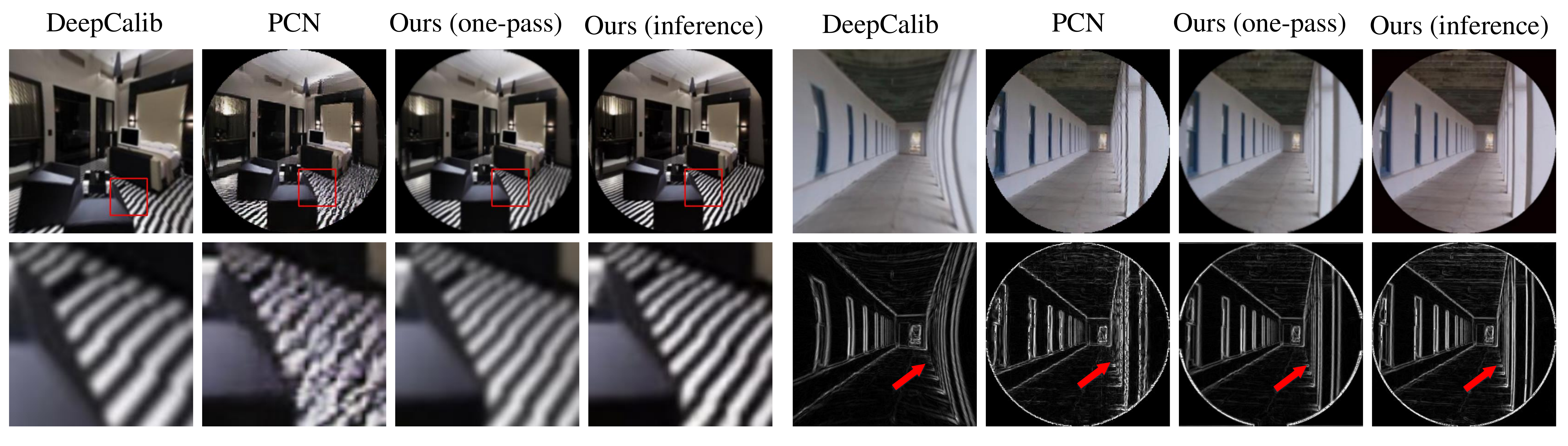}     
  \vspace{-0.3cm}
  \caption{
  \label{special_result}
  \textbf{Additional comparisons on some better performance methods.} We enlarged the local region (marked by red boxes on the left) to compare the image texture. Besides, we highlighted the structural differences (marked by red arrows on the right).}
  \vspace{-0.3cm}
  \end{figure*}

\begin{figure*}[!t]
    \centering
    \includegraphics[scale=.17]{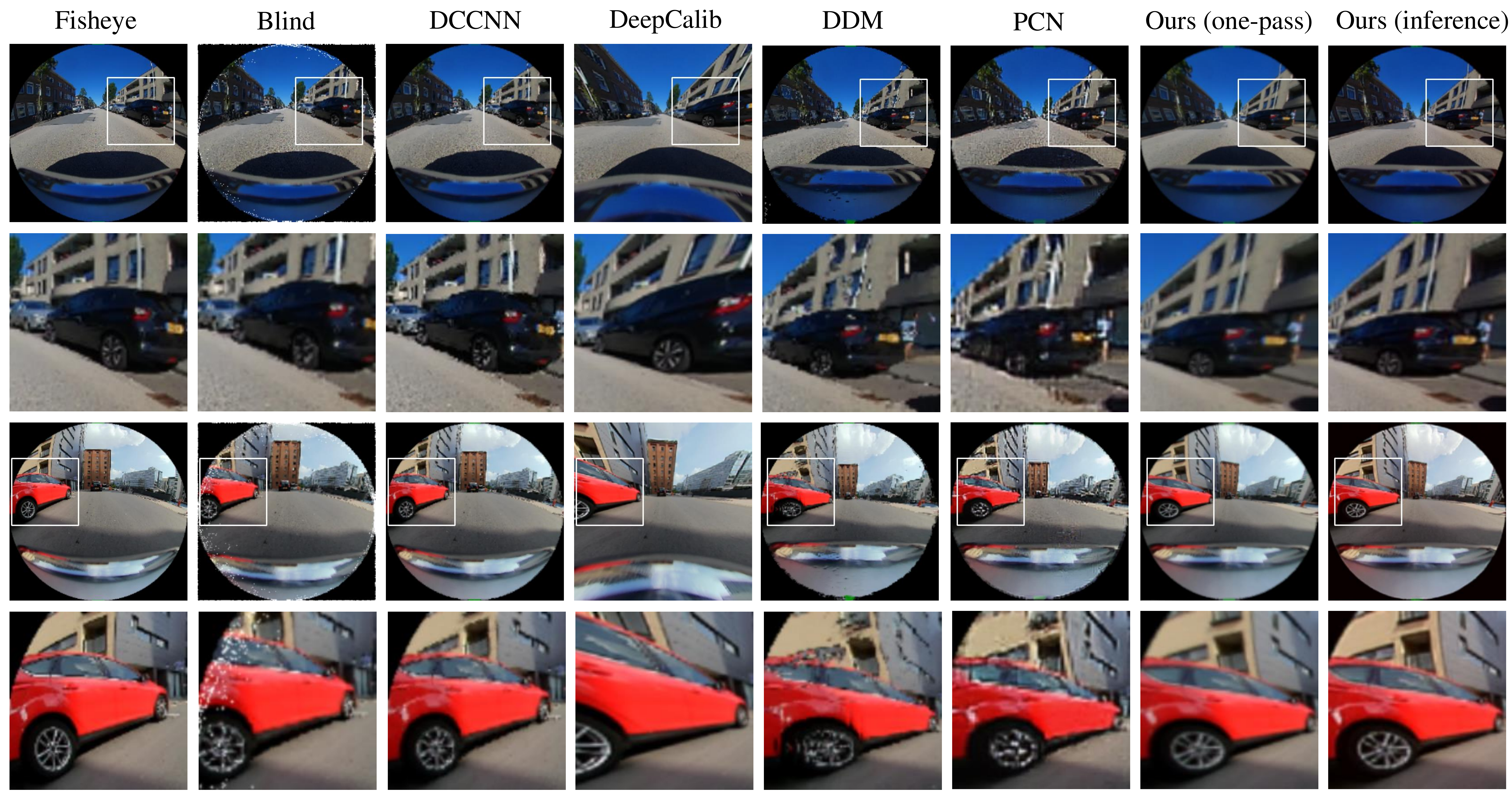}     
    \vspace{-0.2cm}
    \caption{
    \label{real_result}
    \textbf{Visualization results on the real fisheye correction.} We visualize the correction results of the mainstream methods and our methods on the Woodscape dataset \cite{woodscape}, and we enlarge local areas of the results. It is evident that compared with the mainstream methods, our results have a more accurate structure as well as realistic texture.}
    \vspace{-0.2cm}
    \end{figure*}

\begin{table*}[htbp]
  \centering
  \caption{Performance comparison with the state-of-the-art methods.}
  \vspace{-0.2cm}
    \begin{tabular}{lcccccccccc}
    \toprule
    \multicolumn{3}{c}{Comparison} &       & \multicolumn{7}{c}{Metrics} \\
\cmidrule{1-3}\cmidrule{5-11}    Methods &       & \multicolumn{1}{c}{Type} &       & \multicolumn{1}{c}{PSNR $\uparrow$} & \multicolumn{1}{c}{SSIM $\uparrow$ } & \multicolumn{1}{c}{MS-SSIM $\uparrow$} &       & \multicolumn{1}{c}{FID $\downarrow$} & \multicolumn{1}{c}{LPIPS-Alex $\downarrow$} & \multicolumn{1}{c}{LPIPS-Vgg $\downarrow$} \\
\cmidrule{1-1}\cmidrule{3-3}\cmidrule{5-7}\cmidrule{9-11}    Blind \cite{Blind} &       &  Regression     &       &  14.7     & 0.47     &  0.55     &        &  211.3     &  0.434     & 0.427 \\

DCCNN \cite{Rong2016Radial} &       &   Regression    &       &  15.2     &   0.48    &   0.37    &       &  190.8     &   0.289    & 0.345 \\

DeepCalib \cite{DeepCalib} &       &  Regression     &       &  20.8     &   0.69    &   0.77    &       &   69.7    &   0.136    & 0.195 \\
\midrule
DDM \cite{DDM}&       &   Generation    &       &  24.7     &  0.80     &   0.92    &       & 79.5     &   0.142    & 0.238 \\

PCN \cite{PCN}&       &    Generation   &       &  25.1     &   0.82    &  0.92     &       &  65.8     &    0.106   &  0.165\\
\midrule
Ours (one-pass) &      &   Generation    &       &   \textbf{26.0}     &  \textbf{0.85}     &   \textbf{0.95}    &       &  57.8     &   0.149    & 0.123 \\

Ours (inference) &       &   Generation    &       &   24.6    &  0.76     &  0.92     &       &  \textbf{24.9}     &   \textbf{0.061}    & \textbf{0.100} \\
\bottomrule
    \end{tabular}%
  \label{tab_result}%
\vspace{-0.3cm}
\end{table*}%

\section{Experiments}
\subsection{Experiment Setting}
In order to enable the dual diffusion architecture to perceive the similar distribution of real and synthetic fisheye images at the training stage, we need to employ them for training. Since the polynomial model and the division model is similar, we first refer to several previous methods \cite{Xue2019}\cite{DDM}\cite{PCN} and utilize a polynomial model containing four parameters to generate the synthetic fisheye dataset. Our perspective image dataset is the Places2 dataset \cite{Places2}, which contains 10 million perspective images. We randomly selected 44K images (40K for training, 4K for testing). For each perspective image, we randomly set the values of 4 parameters according to the \cite{DDM}\cite{PCN} to generate distortion. As for the real fisheye images, we use the Woodscape dataset \cite{woodscape}, which is the mainstream dataset and comprises over 8K real fisheye images of on-road driving. We randomly selected 8K images for training and 200 for testing. To eliminate the problem of quantitative imbalance between the real and synthetic fisheye images, we perform data augmentation on the real fisheye images. The images sent to the network are resized to $256 \times 256$. Subsequently, we set the initial learning rate and batch size to 1e-4 and 2, respectively, and trained 50 epochs on eight NVIDIA RTX A4000.

\begin{table*}[htbp]
  \footnotesize
  \centering
  \caption{Performance comparison on different architecture.}
  \vspace{-0.2cm}
    \begin{tabular}{lccccccccccccc}
    \toprule
    \multicolumn{1}{c}{\multirow{2}[4]{*}{Architecture}} & \multicolumn{1}{c}{\multirow{2}[4]{*}{Synthetic}} & \multicolumn{1}{c}{\multirow{2}[4]{*}{Real}} & \multicolumn{5}{c}{One-Pass}          &       & \multicolumn{5}{c}{Inference} \\
\cmidrule{4-8}\cmidrule{10-14}          &       &       & \multicolumn{1}{l}{PSNR} & \multicolumn{1}{l}{SSIM} & \multicolumn{1}{l}{MS-SSIM} & \multicolumn{1}{l}{FID} & \multicolumn{1}{l}{LPIPS} &       & \multicolumn{1}{l}{PSNR} & \multicolumn{1}{l}{SSIM} & \multicolumn{1}{l}{MS-SSIM} & \multicolumn{1}{l}{FID} & \multicolumn{1}{l}{LPIPS} \\
    \midrule
    w/ CDM &   \Checkmark    &   \XSolidBrush    &   $-$   &   $-$    &  $-$     &   $-$    &   $-$    &       &  18.6     &   0.55    &   0.72    &  127.9     & 0.191 \\
    w/ CDM+OPN &  \Checkmark     &  \XSolidBrush      &    25.9     &   0.85    &  0.94     &  56.4     &  0.146     &       &   22.1    &   0.71    &   0.84    &  55.1     & 0.101 \\
    w/ DDA &  \Checkmark     &  \Checkmark     &   26.0   &  0.85     &  \textbf{0.95}     &  57.8     &   0.149    &       &  \textbf{24.6}     &   \textbf{0.76}    &  \textbf{0.92}     &  \textbf{24.9}     &  \textbf{0.061}\\
    w/ DDA (w/o EX) &  \Checkmark     &  \Checkmark     &    \textbf{26.2}      &  \textbf{0.86}     &  \textbf{0.95}     &   \textbf{54.3}    &  \textbf{0.139}     &       &  22.05     &  0.70     &  0.83   &  50.6      & 0.090 \\
    \bottomrule
    \end{tabular}%
  \label{tab_ablation}%
  \vspace{-0.4cm}
\end{table*}%

\begin{figure}[!t]
  \centering
  \includegraphics[scale=.163]{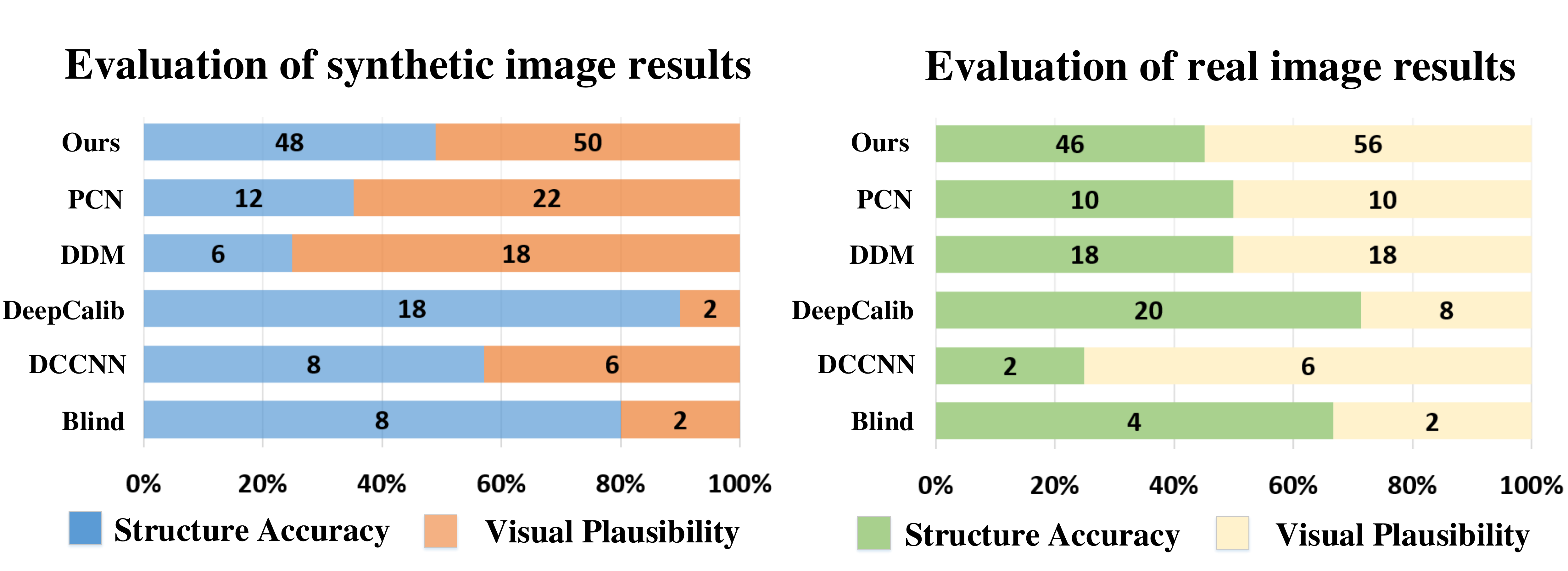}     
  \caption{
  \label{user_study}
  \textbf{Subjective evaluation.} Our synthetic (left) and real (right) results are evaluated more favorably than mainstream methdos in terms of structure accuracy and visual plausibility.}
  \vspace{-0.5cm}
  \end{figure}


\subsection{Subjective and Objective Comparison}
To measure the performance of the methods, we used the same dataset to retrain multiple mainstream correction methods, including Blind \cite{Blind}, DCCNN \cite{Rong2016Radial}, DDM \cite{DDM}, PCN \cite{PCN}. We also compare with DeepCalib \cite{DeepCalib}, which employs a sphere model. However, we cannot retrieve the panorama dataset corresponding to the Places2 dataset \cite{Places2} since DeepCalib uses panorama to synthesize the dataset. Therefore, we employ the pre-trained model directly and crop its output to retain a maximum resemblance to the ground truth. We visualize the correction results of each method and use common metrics, including PSNR, SSIM, FID \cite{FID}, MS-SSIM \cite{MS-SSIM}, LPIPS-Alex \cite{LPIPS}, LPIPS-Vgg \cite{LPIPS}, to quantify the objective performance. 

The subjective and objective results are demonstrated in Figure \ref{result} and Table \ref{tab_result}, respectively. Blind and DCCNN have incomplete corrections due to the simplistic model and the distortion interval. DeepCalib achieves good results by leveraging a more realistic spherical model. However, DDM and PCN achieved more substantial progress with network improvements. Their recursive correction led to better subjective and objective results. In contrast, our method applies a novel diffusion model to improve quality. Our network combines the benefits of GANs with the diffusion model, thus providing two correction schemes at the same time. Particularly, our one-pass correction and inference correction results achieve optimal objective performance in terms of distortion metrics (PSNR, SSIM, MS-SSIM) and the perception metric (FID, LPIPS-Alex, LPIPS-Vgg), respectively. Furthermore, in terms of subjective vision, our inference results outperform the compared schemes by restoring the maximum sharpness and accomplishing rectification, which is demonstrated in Figure \ref{special_result}. To further evaluate the subjective performance, we recruited 20 volunteers to select the best correction results of 100 randomly selected fisheye images. Figure \ref{user_study} demonstrates that our one-pass results obtain greater access in distortion correction, while our inference results increase in quality improvement.

\subsection{Comparison on Real Fisheye Image Correction}
To prove that our method implements generalized correction, we visualize the correction results of real fisheye images, as shown in Figure \ref{real_result}. It can be seen that Blind and DCCNN have incomplete corrections. DDM and PCN can achieve correction for real fisheye images, but the results have obvious artifacts. Because of domain discrepancy, the model trained on the synthetic datasets can only effectively correct the synthetic fisheye images. DeepCalib performs well because the sphere model is more consistent with the real fisheye distribution. However, it crops the image boundaries and loses a lot of information. In contrast, the one-pass correction and inference correction in our method both achieve good correction. Our results retain the whole image boundary, and the visual impression of our inference scheme has significant clarity. Therefore, our correction effect on the real fisheye image outperforms all previous methods. The subjective evaluation in Figure \ref{user_study} also proves that our results are more likely to be accepted.

\begin{figure}[!t]
  \centering
  \includegraphics[scale=.185]{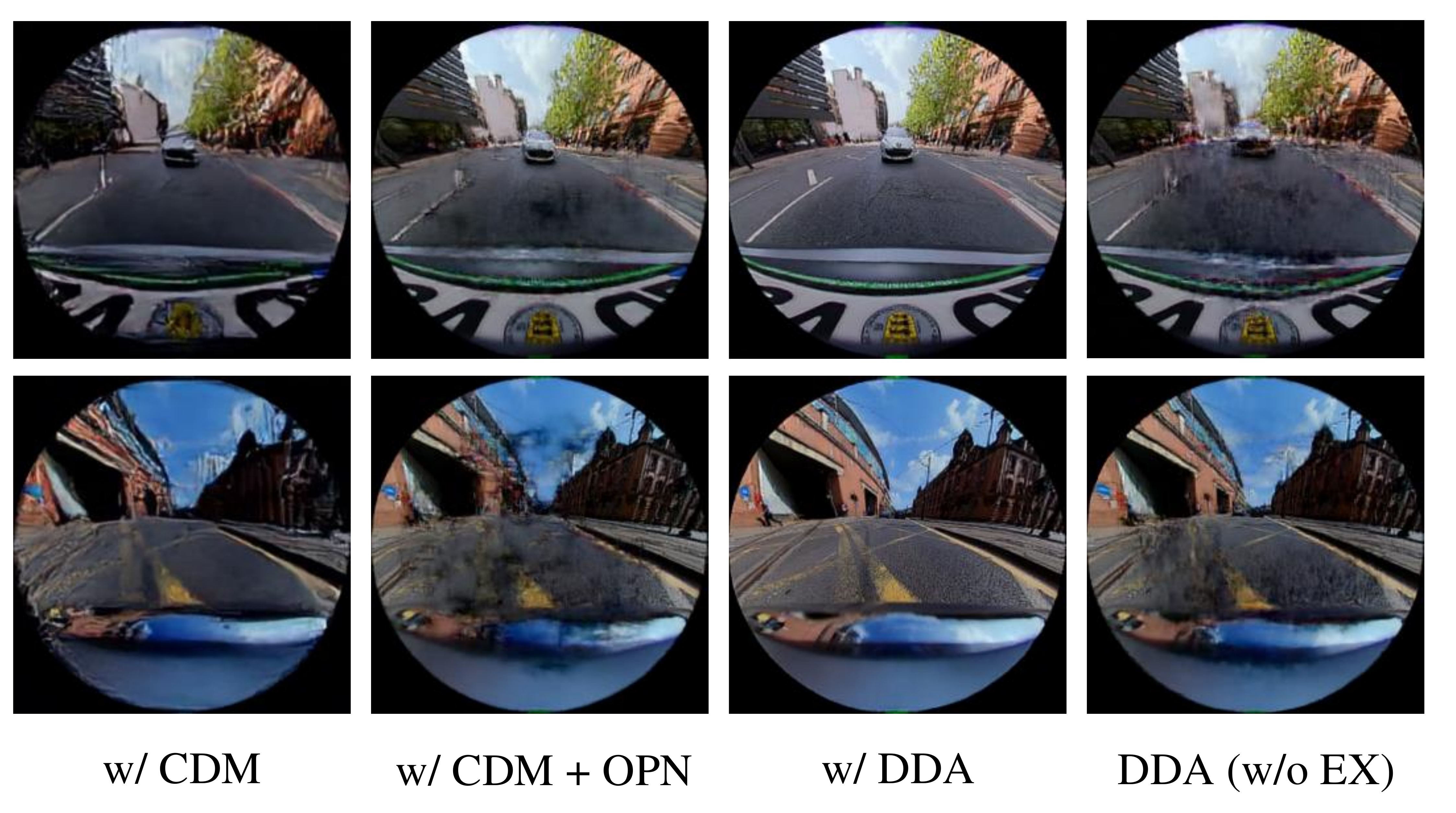}     
  \vspace{-0.7cm}
  \caption{
  \label{ab_st_result}
  Visualization results of different architectures.}
  \vspace{-0.5cm}
  \end{figure}

\subsection{Ablation Study}
The unconditional diffusion module (UDM), conditional diffusion module (CDM), and OPN are the major components of our dual-diffusion architecture. To verify the effectiveness of each module, we begin with the original conditional diffusion module, then add each module and evaluate the improvement. Ablation study results are shown in Figure \ref{ab_st_result} and Table \ref{tab_ablation}. First, we use a naive conditional diffusion model (w/ CDM) with the original synthetic fisheye as the condition to correct our fisheye image. It can be seen that the performance is poor, and the correction results exhibit significant blurring. Subsequently, we increase OPN (w/ CDM + OPN). It brings a qualitative improvement in the synthetic correction results, but the correction of real fisheye images is unsatisfactory. Subsequently, we further increase the unconditional diffusion module (w/ DDA) and introduce the unlabeled real fisheye image for training. The network achieves satisfactory performance in the synthetic and real fisheye images. Furthermore, to demonstrate the importance of interaction between two diffusion modules, we remove the information that they exchange (w/o EX). The drop in real correction performance indicates that the correction principles learned from the synthetic images cannot be applied effectively to the real images. As a result, although the performance on synthetic images is slightly better, the effect of the real image is significantly decreased.


\subsection{Limitaion Discussion}
Most previous methods effectively correct synthetic fisheye images but fail on real fisheye images. Therefore, we designed a dual diffusion architecture that can transform synthetic and real fisheye images into a consistent noise distribution. As a result, cross-domain correction for real fisheye is accomplished. It can be seen in Figure \ref{real_result} that our method significantly improved the real correction performance compared with the mainstream methods.

However, our method still has drawbacks. The enormous border distortion cannot be totally erased by our network since the real fisheye image is unlabeled. Nonetheless, our network still significantly reduces the distortion and produces high-quality results, which is a worthwhile exploration. Completely eliminating the huge boundary distortion is a more difficult challenge. We will further investigate it in future work.

\section{Conclusion}
In this paper, we propose a novel dual diffusion architecture to address the domain discrepancy problem. It can simultaneously use paired synthetic fisheye images and unlabeled real fisheye images for training. Different from the previous diffusion methods, we attempt to introduce a one-pass network in the conditional diffusion module to provide new reasonable guidance for the diffusion module. The one-pass network can achieve unsupervised training and provide an optional quick correction scheme in the test stage. Our dual diffusion architecture combines conditional and unconditional diffusion modules together. By supervising the same noise intensity, we can transform two inconsistent distributions into a consistent noise distribution and achieve cross-domain correction. Numerous experiments demonstrate that the one-pass and inference results in our methods both outperform the competition.

{\small
\bibliographystyle{unsrt}
\bibliographystyle{ieee_fullname}
\bibliography{ref}
}

\end{document}